\documentclass[conference]{IEEEtran}
\IEEEoverridecommandlockouts
\usepackage[section]{placeins}
\usepackage{float}

\usepackage{cite}
\usepackage{amsmath,amssymb,amsfonts}
\usepackage{algpseudocode}
\usepackage{algorithm}
\usepackage{graphicx}
\usepackage[font=normalsize,labelfont=bf]{caption}
\usepackage{textcomp}
\usepackage{xcolor}

\usepackage[compact]{titlesec}
\titlespacing*{\section}{0pt}{0.7em}{0.4em}
\titlespacing*{\subsection}{0pt}{0.5em}{0.25em}
\titlespacing*{\paragraph}{0pt}{0.25em}{0.25em}

\def\BibTeX{{\rm B\kern-.05em{\sc i\kern-.025em b}\kern-.08em
\kern-.1667em\lower.7ex\hbox{E}\kern-.125emX}}

\begin{document}

\title{Predictive Modeling in AUV Navigation: A Perspective from Kalman Filtering 
}

\author{
\IEEEauthorblockN{
\textsuperscript{*}Zizhan Tang\textsuperscript{1},
\textsuperscript{*}Yao Liu\textsuperscript{1}, and
Jessica Liu\textsuperscript{2}
}
\IEEEauthorblockA{
\textsuperscript{1}Department of Electrical and Computer Engineering, University of Southern California, Los Angeles, CA, USA\\
\textsuperscript{2}Department of Computer and Information Science, University of Pennsylvania, Philadelphia, PA, USA\\
Email: \{ztang, jliu3548\}@usc.edu, liujiexi@seas.upenn.edu
}
}
\maketitle

\begin{abstract}
We present a safety-oriented framework for autonomous underwater vehicles (AUVs) that improves localization accuracy, enhances trajectory prediction, and supports efficient search operations during communication loss. Acoustic signals emitted by the AUV are detected by a network of fixed buoys, which compute Time-Difference-of-Arrival (TDOA) range-difference measurements serving as position observations. These observations are subsequently fused with a Kalman-based prediction model to obtain continuous, noise-robust state estimates. The combined method achieves significantly better localization precision and trajectory stability than TDOA-only baselines.

Beyond real-time tracking, our framework offers targeted search-and-recovery capability by predicting post-disconnection motion and explicitly modeling uncertainty growth. The search module differentiates between continued navigation and propulsion failure, allowing search resources to be deployed toward the most probable recovery region. Our framework fuses multi-buoy acoustic data with Kalman filtering and uncertainty propagation to maintain navigation accuracy and yield robust search-region definitions during communication loss.
\end{abstract}

\begin{IEEEkeywords}
Predictive Trajectory Models and Motion Forecasting; Continuous Localization Solutions; Sensor Fusion for Accurate 
\end{IEEEkeywords}

\section{Introduction}

Autonomous Underwater Vehicles (AUVs) frequently operate in environments where GPS is unavailable and acoustic communication is intermittent or prone to loss. As a result, reliable localization becomes especially challenging when the vehicle unexpectedly loses communication with surface buoys or support vessels. In this work, we address the problem of \textit{search and recovery under communication loss}, where the goal is to estimate the AUV’s most probable location after disconnection and construct uncertainty-driven search regions for recovery operations. We consider two practical operating assumptions: (1) the AUV may continue following its planned trajectory after communication loss with no mechanical failures, and (2) the AUV may lose propulsion at the moment of disconnection, resulting in passive drift driven by environmental disturbances. These assumptions define two motion models that guide our predictive estimation and search framework.

We first describe the localization framework used to obtain the AUV’s last reliable state prior to communication loss. 
Classical TDOA analyses such as Gezici’s survey~\cite{gezici2007} establish fundamental estimation limits for range-difference positioning, characterizing achievable accuracy under realistic noise. More recent underwater-specific formulations leverage optimization to improve robustness under multipath and noise, including the MM-based TDOA estimator of Li et al.~\cite{li2020mmtdoa}, which outperforms classical least-squares approaches under high measurement uncertainty. In practical underwater systems, however, acoustic ranges are corrupted by multipath, timing jitter, and low-bandwidth communication effects, as characterized in Stojanovic’s work on underwater acoustics~\cite{stojanovic2009underwater}. These challenges motivate the use of sequential estimation techniques that smooth noisy observations and enforce temporal consistency. Modern underwater navigation frameworks fuse acoustic measurements with vehicle dynamics through a variety of estimation techniques, including invariant-EKF inertial–acoustic fusion~\cite{mangelson2021inekf}, nonlinear observer designs for current-driven motion~\cite{mangelson2021inekf}, and graph-optimized INS/USBL/DVL integration~\cite{li2023graphfusion}.
Furthermore, Braginsky et al.~\cite{braginsky2016tracking} demonstrate an ASV–AUV ranging architecture in which periodic acoustic interrogations from a surface vehicle provide intermittent range updates that reduce drift during normal operation.
Although such approaches improve robustness during normal operation, they fundamentally rely on the availability of continuous or intermittent measurements to remain observable. As summarized in the survey by Kinsey et al.~\cite{kinsey2006navigation}, modern underwater navigation frameworks combine INS, DVL, and acoustic measurements to maintain bounded-error positioning during normal operation. Recent work such as Tang et al.~\cite{tang2024acoustic} demonstrates that inertial uncertainty grows rapidly when acoustic updates become sparse, reinforcing the need for predictive uncertainty modeling when external observations are unavailable.

Search‑and‑rescue (SAR) modeling provides additional foundations for lost‑target recovery, including probabilistic drift forecasting and search‑effort allocation frameworks developed for maritime operations.
Classical Bayesian search theory~\cite{stone1981search, stone1975optimal} provides principled strategies for allocating search effort, but presumes a known prior distribution over the target location. Drift-only maritime SAR models, surveyed by Selezin~\cite{selezensearch}, model only passive drift under wind and ocean currents and cannot reconstruct pre‑disconnection motion or account for AUV‑specific failure modes.

Thus, estimation methods characterize vehicle behavior while measurements are available, and SAR models capture uncertainty growth once measurements are lost, but neither connects the last known state to the post-loss uncertainty evolution. This gap motivates the unified pre‑ and post‑disconnection framework proposed in this work.

Our work connects acoustic sensing, sequential Bayesian estimation, and uncertainty-driven search planning into a single, cohesive pipeline designed for search and recovery protocol during communication loss. Our contributions are as follows:

\begin{itemize}
    \item A multi-buoy acoustic localization module using a Chan-based closed-form TDOA estimator, providing geometrically consistent 3D position estimates without iterative optimization.
    \item A sequential Kalman-filter prediction layer that fuses acoustic measurements with vehicle dynamics, yielding smooth, real-time trajectory estimates robust to noisy sensing.
    \item A post-disconnection prediction model that explicitly propagates uncertainty over time, distinguishing between continued motion and propulsion failure and producing principled, time-varying search regions.
    \item A unified safety-driven framework that links sensing, estimation, and adaptive search-region construction, enabling targeted recovery operations after communication loss.
\end{itemize}
The remainder of this paper is organized as follows: Section~\ref{sec:preliminaries} introduces the system model and preliminary formulations. Section~\ref{sec:methodology} presents the localization and prediction methods based on TDOA and Kalman filtering. Section~\ref{sec:search_model} describes the search model for disconnected AUVs. Simulation results are provided in Section~\ref{sec:simulation}, followed by conclusions in Section~\ref{sec:conclusion}.

\section{Preliminaries}
\label{sec:preliminaries}

We consider an underwater submersible whose position cannot be directly obtained via GPS due to severe signal attenuation. 
The submersible is equipped with an acoustic transmitter that periodically emits signals received by multiple buoys deployed at known locations. 
The \textit{Time Difference of Arrival} (TDOA) of these acoustic signals provides a measurement of the submersible’s position, which is then fused with inertial dynamics through a Kalman filter for sequential state estimation under process and observation noise.

The theoretical foundation for the estimation framework builds upon the classical linear filtering approach proposed by Kalman~\cite{kalman1960} and the hyperbolic multilateration solution introduced by Chan and Ho~\cite{chan1994}, which together form the basis for integrating acoustic localization with predictive state estimation in underwater environments.

\subsection{TDOA-Based Localization}

Let the true position of the submersible at time step $t$ be denoted as
\[
\mathbf{p}_t = [x_t,\, y_t,\, z_t]^{\top}.
\]
A set of $N$ surface buoys are placed at fixed and known coordinates
\[
\mathbf{s}_i = [x_i,\, y_i,\, z_i]^{\top}, \quad i = 0, \dots, N-1,
\]
where $s_0$ serves as the reference station. 
Each buoy records the signal arrival time $t_i$, and the TDOA between buoy $i$ and the reference buoy is defined by
\[
\Delta t_i = t_i - t_0.
\]
\begin{equation}
c \, \Delta t_i = \|\mathbf{p}_t - \mathbf{s}_i\|_2 - \|\mathbf{p}_t - \mathbf{s}_0\|_2 + \varepsilon_i,
\label{eq:tdoa}
\end{equation}
where $\varepsilon_i \sim \mathcal{N}(0, R_i)$ models measurement noise. 

\begin{figure}[!htbp]
    \centering
    \includegraphics[width=0.8\linewidth]{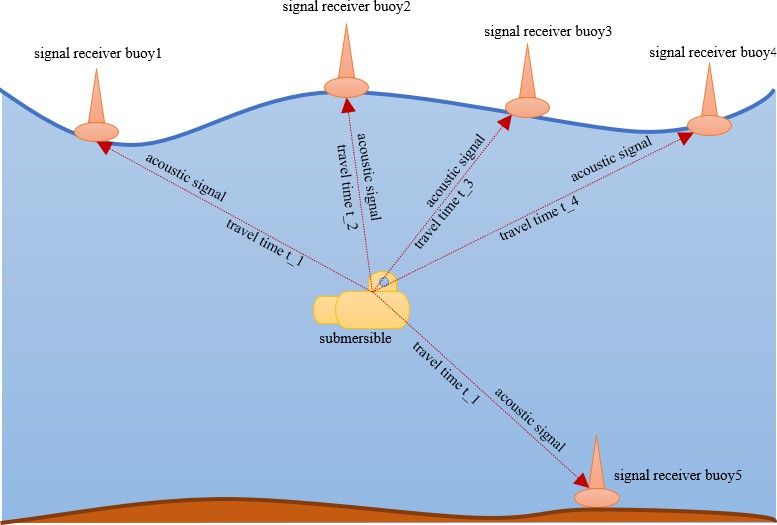}
    \caption{Illustration of the TDOA-based localization system. The submersible emits acoustic signals received by multiple buoys at different times, from which the travel time differences are used to estimate its position.}
    \label{fig:tdoa_system}
\end{figure}

Linearizing \eqref{eq:tdoa} around a nominal position $\mathbf{p}_0$ yields the linear system
\begin{equation}
A_t \, \mathbf{p}_t = \mathbf{b}_t + \boldsymbol{\varepsilon}_t,
\label{eq:tdoa_linear}
\end{equation}
where $A_t \in \mathbb{R}^{(N-1)\times 3}$ and $\mathbf{b}_t \in \mathbb{R}^{(N-1)}$ are determined by spatial geometry and partial derivatives of \eqref{eq:tdoa}. 
The least-squares solution provides the observed position estimate
\begin{equation}
\hat{\mathbf{p}}_t = (A_t^{\top} A_t)^{-1} A_t^{\top} \mathbf{b}_t.
\label{eq:tdoa_estimate}
\end{equation}
This estimated position $\hat{\mathbf{p}}_t$ serves as the measurement input to the Kalman filter described next.
Eq.~(2) provides an initial linearized TDOA estimate, which we refine using Chan’s closed-form multilateration ~\cite{chan1994} for improved geometric accuracy.

\subsection{Kalman Filtering Model}

The submersible’s motion is modeled as a discrete-time linear Gaussian system:
\begin{equation}
\mathbf{x}_{t+1} = F\,\mathbf{x}_t + G\,\mathbf{u}_t + \mathbf{w}_t,
\qquad
\mathbf{z}_t = H\,\mathbf{x}_t + \mathbf{v}_t,
\label{eq:kalman_model}
\end{equation}
where
$\mathbf{x}_t = [x_t,\, y_t,\, z_t,\, v_{x,t},\, v_{y,t},\, v_{z,t}]^{\top}$ 
is the state vector,
$\mathbf{u}_t = [a_{x,t},\, a_{y,t},\, a_{z,t}]^{\top}$ 
is the control input (engine and current accelerations),
and $\mathbf{z}_t = \hat{\mathbf{p}}_t$ 
is the observation from the TDOA subsystem.
The process and observation noise terms 
$\mathbf{w}_t \sim \mathcal{N}(0, Q)$ and 
$\mathbf{v}_t \sim \mathcal{N}(0, R)$ 
are zero-mean and mutually independent.

For a constant-acceleration kinematic model, the state transition and observation matrices are given by
\[
F =
\begin{bmatrix}
I_3 & T I_3 \\
0_3 & I_3
\end{bmatrix},
\quad
G =
\begin{bmatrix}
\frac{1}{2} T^2 I_3 \\[2pt]
T I_3
\end{bmatrix},
\quad
H = [I_3 \;\; 0_3],
\]
where $T$ is the sampling interval.

Given the prior estimate $\hat{\mathbf{x}}_{t|t-1}$ and covariance $P_{t|t-1}$, the Kalman recursion proceeds as:

\textbf{Prediction Step:}
\begin{subequations}\label{eq:kalman_predict}
\begin{align}
\mathbf{x}_{t|t-1} &= F\,\hat{\mathbf{x}}_{t-1|t-1} + G\,\mathbf{u}_{t-1}, \tag{5a}\\[4pt]
P_{t|t-1} &= F\,P_{t-1|t-1}\,F^{\top} + Q. \tag{5b}
\end{align}
\end{subequations}

\textbf{Update Step:}
\begin{subequations}\label{eq:kalman_update}
\begin{align}
K_t &= P_{t|t-1}\,H^{\top}\!\left(H\,P_{t|t-1}\,H^{\top} + R\right)^{-1}, \tag{6a}\\[6pt]
\hat{\mathbf{x}}_{t|t} &= \mathbf{x}_{t|t-1} + K_t\!\left(\mathbf{z}_t - H\,\mathbf{x}_{t|t-1}\right), \tag{6b}\\[4pt]
P_{t|t} &= (I - K_t H)\,P_{t|t-1}. \tag{6c}
\end{align}
\end{subequations}

Equations (4) together with the Kalman Prediction and Update equations (5a)–(5b) and (6a)–(6c) define the recursive Bayesian estimator that fuses acoustic observations from TDOA with the dynamic state model of the submersible, yielding minimum-variance estimates of its position and velocity under Gaussian uncertainty assumptions.

\paragraph{Kalman Operators for Succinctness.}
For succinctness, we define the \textit{Kalman Prediction} and \textit{Kalman Update} operators, denoted $\mathrm{Predict}(\cdot)$ and $\mathrm{Update}(\cdot)$, which compactly represent the full recursion in Eqs.~(5a)–(5b) and Eqs.~(6a)–(6c), respectively. These operators are used in Algorithm~\ref{alg:localization-kalman} to streamline notation while preserving the exact mathematical meaning of the standard Kalman filter update.

\section{Methodology}
\label{sec:methodology}

In this section, we detail the mathematical formulation of the localization process based on the \textit{Time Difference of Arrival} (TDOA) technique and the subsequent derivation of the submersible’s position. We employ a closed-form solution following the framework of Chan’s algorithm~\cite{chan1994}, which provides a computationally efficient and geometrically consistent approach to multilateration in three-dimensional space.

\subsection{TDOA Localization Derivation}
\label{sec:tdoa_localization}

To derive the position of the submersible from Time Difference of Arrival (TDOA) measurements, we follow the closed-form multilateration framework proposed by Chan~\cite{chan1994}. The goal is to estimate the position vector $\mathbf{p} = [x,\, y,\, z]^{\top}$ using acoustic arrival times at a set of fixed receivers.

\paragraph{System Setup.}
Consider one reference receiver $\mathbf{s}_0 = [x_0,\, y_0,\, z_0]^{\top}$ and four other receivers $\mathbf{s}_i = [x_i,\, y_i,\, z_i]^{\top}$ for $i = 1,2,3,4$. Each receiver records the signal arrival time $t_i$, and the TDOA relative to the reference receiver is
\[
\Delta t_i = t_i - t_0, \qquad i = 1,2,3,4.
\]
Assuming a constant acoustic propagation speed $c$, the difference in signal travel distances is
\[
c\,\Delta t_i = d_i - r_0,
\qquad
d_i = \|\mathbf{p} - \mathbf{s}_i\|_2, \quad
r_0 = \|\mathbf{p} - \mathbf{s}_0\|_2.
\]

\paragraph{Linearization.}
Squaring both sides of~\eqref{eq:tdoa} and expanding the range expressions yield a set of nonlinear equations. Introducing $k_i = x_i^2 + y_i^2 + z_i^2$ and rearranging terms with respect to $\mathbf{p}$ gives the linear form:
\begin{equation}
A\,\mathbf{p} = r_0\,\mathbf{C} + \mathbf{D},
\label{eq:chan_linear}
\end{equation}
where
\[
A =
\begin{bmatrix}
x_1 - x_0 & y_1 - y_0 & z_1 - z_0\\
x_2 - x_0 & y_2 - y_0 & z_2 - z_0\\
x_3 - x_0 & y_3 - y_0 & z_3 - z_0\\
x_4 - x_0 & y_4 - y_0 & z_4 - z_0
\end{bmatrix}, \quad
\mathbf{C} =
\begin{bmatrix}
\Delta t_1\\ \Delta t_2\\ \Delta t_3\\ \Delta t_4
\end{bmatrix},
\]
%
\[
\mathbf{D} = \frac{1}{2}
\begin{bmatrix}
  k_1 - k_0 - c^2(\Delta t_1)^2 \\[4pt]
  k_2 - k_0 - c^2(\Delta t_2)^2 \\[4pt]
  k_3 - k_0 - c^2(\Delta t_3)^2 \\[4pt]
  k_4 - k_0 - c^2(\Delta t_4)^2
\end{bmatrix}.
\]

\paragraph{Closed-Form Solution.}
Equation~\eqref{eq:chan_linear} defines a linear system in the unknowns $(x, y, z)$ and $r_0$. Following Chan’s algorithm, the position vector is expressed as a linear combination of the two least-squares solutions to $A\mathbf{x}=\mathbf{C}$ and $A\mathbf{x}=\mathbf{D}$:
\begin{equation}
\mathbf{p} = \mathbf{a}\,r_0 + \mathbf{b},
\qquad
\mathbf{a} = A^{\dagger}\mathbf{C}, \quad
\mathbf{b} = A^{\dagger}\mathbf{D},
\label{eq:chan_solution}
\end{equation}
where $A^{\dagger}$ denotes the Moore–Penrose pseudoinverse.

\paragraph{Quadratic Solution for $r_0$.}
Substituting~\eqref{eq:chan_solution} into the distance constraint $r_0 = \|\mathbf{p} - \mathbf{s}_0\|_2$ leads to a quadratic equation in $r_0$:
\[
\tilde{A}r_0^2 + \tilde{B}r_0 + \tilde{C} = 0,
\]
where the coefficients are
\[
\begin{aligned}
\tilde{A} &= a_1^2 + a_2^2 + a_3^2,\\
\tilde{B} &= 2(a_1b_1 + a_2b_2 + a_3b_3 - a_1x_0 - a_2y_0 - a_3z_0),\\
\tilde{C} &= (x_0 - b_1)^2 + (y_0 - b_2)^2 + (z_0 - b_3)^2.
\end{aligned}
\]
The valid physical solution corresponds to the positive root:
\[
r_0 = \frac{-\tilde{B} + \sqrt{\tilde{B}^2 - 4\tilde{A}\tilde{C}}}{2\tilde{A}}.
\]
Finally, substituting $r_0$ into~\eqref{eq:chan_solution} yields the 3D position estimate $\mathbf{p} = [x, y, z]^{\top}$.

\paragraph{Remarks.}
This closed-form formulation avoids iterative nonlinear optimization and provides robust geometric consistency for underwater localization. In practice, additional buoys can be incorporated to form an overdetermined system, solved via least squares to improve noise resilience and positioning accuracy.

Compared with iterative least-squares solvers, Chan’s closed-form TDOA algorithm offers significant computational efficiency, requiring only matrix inversion and a quadratic evaluation, which makes it well suited for real-time embedded AUV applications.

\label{sec:kalman}
\subsection{Kalman Filtering for Sequential Prediction}
The acoustic TDOA subsystem described above provides noisy position observations $\mathbf{z}_t = \hat{\mathbf{p}}_t$ at discrete time steps. 
To obtain smooth and temporally consistent state estimates, we employ a Kalman filter that fuses these measurements with a dynamic motion model of the submersible. 
This subsection outlines the full formulation, including initialization, recursive estimation, and motion equations consistent with the physical acceleration model.

\paragraph{Initialization.}
\noindent
The measurement noise covariance $R$ represents uncertainty in the range‑difference measurements produced by the TDOA subsystem. In practice, this noise reflects timing jitter, hydrophone resolution, and small variations in acoustic propagation. Without loss of generality, we model these effects as independent Gaussian errors across buoys, resulting in a diagonal $R$ matrix whose entries capture the variance of the TDOA measurement noise. This formulation allows the Kalman filter to appropriately weight the relative trust between model predictions and acoustic observations.

The process noise covariance $Q$ models uncertainty in the submersible’s motion dynamics, such as unmodeled accelerations or disturbances from ocean currents. Its magnitude and structure are chosen based on prior knowledge of the environment and vehicle behavior, and can be tuned to reflect different operating conditions.

\paragraph{Kalman Recursion.}
The sequential state estimation follows the standard prediction and measurement-update equations already defined in Section~\ref{sec:preliminaries}, specifically Eqs.~(5a)–(5b) and Eqs.~(6a)–(6c). These equations govern the propagation of the prior state, the incorporation of acoustic measurements, and the update of the state covariance without rewriting them here.

\paragraph{State-Space Formulation.}
The submersible’s motion is modeled as a constant-velocity linear system with acceleration input:
\[
\mathbf{x}_{t+1} = F\,\mathbf{x}_t + G\,\mathbf{u}_t + \mathbf{w}_t, \qquad 
\mathbf{z}_t = H\,\mathbf{x}_t + \mathbf{v}_t,
\]
where $\mathbf{x}_t = [x_t,\, y_t,\, z_t,\, v_{x,t},\, v_{y,t},\, v_{z,t}]^{\top}$ represents position and velocity states, and $\mathbf{u}_t$ encodes the engine acceleration and ocean-current disturbances.
The transition and observation matrices follow the standard constant-velocity model with sampling interval $T$:
\[
F =
\begin{bmatrix}
I_3 & T I_3 \\[2pt]
0_3 & I_3
\end{bmatrix}, \qquad
G =
\begin{bmatrix}
\frac{T^2}{2} I_3 \\[2pt]
T I_3
\end{bmatrix}, \qquad
H = [I_3 \;\; 0_3].
\]

\paragraph{Physical Interpretation.}
This formulation captures the coupled evolution of position and velocity driven by acceleration inputs from both propulsion and environmental currents. 
The Kalman filter adaptively fuses TDOA-derived position measurements with model predictions, yielding minimum-variance estimates suitable for real-time underwater tracking.

\FloatBarrier
\begin{figure}[h]
    \centering
    \includegraphics[width=0.8\linewidth]{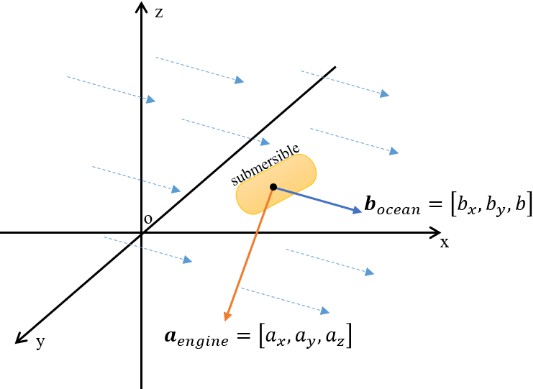}
    \caption{Mechanical interpretation of the submersible motion under combined engine and ocean-current accelerations.}
    \label{fig:kalman_dynamics}
\end{figure}

\section{Search Model}
\label{sec:search_model}

Following the predictive modeling phase, we analyze the search and recovery framework designed for cases in which the submersible loses communication with the surface buoys or host vessel. 
After disconnection, no new observations are received, and the Kalman filter propagates the last state estimate forward to predict the location. 
This prediction guides rescue deployment and defines the most probable recovery region.

\vspace{-0.5em}
\subsection{Disconnection Condition}
A disconnection event is defined as the moment when no valid acoustic signal is detected within a time interval exceeding the nominal sampling period. 
At this instant, the most recent filtered state $\hat{\mathbf{x}}_{t|t}$ and its covariance $P_{t|t}$ are stored as initial conditions for autonomous prediction:
\[
\mathbf{x}_{t_0} = \hat{\mathbf{x}}_{t|t}, \quad P_{t_0} = P_{t|t}.
\]
From that point onward, the system state evolves according to the same discrete‐time dynamics defined in Section~\ref{sec:kalman}. 
The process noise covariance $Q$ governs the rate at which uncertainty increases over time, forming an expanding search radius as prediction continues.

\begin{figure}[!htbp]
    \centering
    \includegraphics[width=0.70\linewidth]{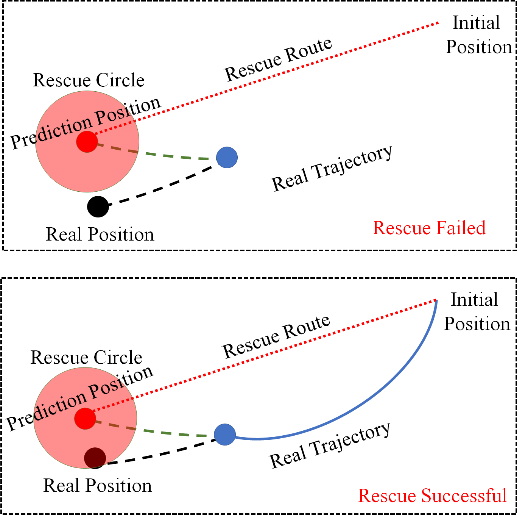}
    \caption{Illustration of the search process under communication loss. 
    The predicted trajectory (red) is generated by the Kalman filter, while the blue curve denotes the true motion. 
    The shaded circles illustrate the evolving search regions under different operating conditions.}
    \label{fig:search_process}
\end{figure}

\vspace{-0.5em}
\subsection{Scenario 1: No Mechanical Failures (Predetermined Navigation)}
This scenario represents the case in which the submersible loses communication but continues normal operation, experiencing no mechanical issues apart from the failure of the communication link. 
The vehicle maintains its predetermined navigation plan—for example, following a sightseeing route along the seabed of the Ionian Sea—where velocity and engine acceleration at each time step are predefined.

The last available observation before disconnection is processed through the Kalman filter to compute the posterior optimal estimate of the submersible’s last known position. 
From this state, the prediction model extrapolates the next position using deterministic dynamics:
\begin{equation}
\mathbf{x}_{t+1} = F\,\mathbf{x}_t + G\,\mathbf{u}_t,
\label{eq:search_predicted}
\end{equation}
where $\mathbf{u}_t$ denotes the control input determined by the propulsion command at the last connected moment. 
Because no new measurements are available, the filter performs only the prediction step; the prior estimate becomes the next posterior. 
This recursive propagation continues along the planned trajectory until reconnection or recovery.

Within several minutes after disconnection, the prediction error remains small—typically within a few meters—since the system follows a known, deterministic path. 
The predicted positions, together with the corresponding uncertainty ellipses derived from $P_t$, define high‐confidence search zones that rescue teams can prioritize.

\vspace{-0.5em}
\subsection{Scenario 2: Loss of Propulsion}
In this scenario, the submersible experiences a mechanical failure such as power loss or fuel depletion immediately before or during disconnection. 
Although the last transmitted state includes the final measured velocity and acceleration, propulsion ceases, and the submersible’s subsequent motion is governed by environmental disturbances such as ocean currents.

The post‐disconnection dynamics are modeled as a random walk process:
\begin{equation}
\mathbf{x}_{t+1} = F\,\mathbf{x}_t + \mathbf{w}_t, 
\qquad \mathbf{w}_t \sim \mathcal{N}(0, Q),
\label{eq:search_randomwalk}
\end{equation}
where $\mathbf{w}_t$ captures random acceleration induced by ocean currents. 
The covariance $P_t$ expands progressively as the system propagates without corrective measurements, defining a time‐varying uncertainty ellipsoid that enlarges the search region.

Although deterministic prediction is no longer reliable, the probabilistic framework still provides a spatial confidence map from which the rescue operation can be planned. 
The predicted mean $\hat{\mathbf{x}}_{t+k}$ serves as the center of the search area, while the corresponding covariance determines its extent. 
Over time, the search radius grows approximately with $\sqrt{\mathrm{tr}(P_{t+k})}$, where $\mathrm{tr}(\cdot)$ denotes the matrix trace (the sum of diagonal elements), quantifying the trade‐off between elapsed time and recovery probability.

Both scenarios demonstrate that the Kalman‐based search model provides a principled mechanism for estimating post‐disconnection trajectories. 
By leveraging the last known motion state, it enables targeted, confidence‐driven search operations that balance coverage efficiency with prediction uncertainty.

\FloatBarrier
\begin{algorithm}[h]
\caption{Real-Time Acoustic Localization and Kalman Filtering Pipeline}
\label{alg:localization-kalman}
\begin{algorithmic}[1]

\Require Buoy positions $\mathcal{B}$; acoustic speed $c$; sampling interval $T$; Kalman model $(F,G,H,Q,R)$  
\Ensure Filtered state estimate $\hat{\mathbf{x}}_{t|t}$

\State Initialize Kalman filter with $\hat{\mathbf{x}}_{0|0}$ and $P_{0|0}$
\State last\_packet\_time $\leftarrow 0$

\While{acoustic\_packets\_received}

    \If{acoustic packets received}
        \State Compute TDOA: $\Delta t_i = t_i - t_0$
        \State Compute Chan multilateration estimate $\hat{\mathbf{p}}_t$
        \State last\_packet\_time $\leftarrow t$
    \EndIf

    \State \textbf{Kalman Prediction Step}
    \State $(\mathbf{x}_{t|t-1},\, P_{t|t-1}) \gets \mathrm{Predict}(\hat{\mathbf{x}}_{t-1|t-1},\, P_{t-1|t-1},\, \mathbf{u}_{t-1})$

    \If{acoustic packets received}
        \State \textbf{Kalman Update Step}
        \State $(\hat{\mathbf{x}}_{t|t},\, P_{t|t}) \gets \mathrm{Update}(\mathbf{x}_{t|t-1},\, P_{t|t-1},\, \hat{\mathbf{p}}_t)$
    \Else
        \State \textbf{No update — continue prediction}
        \State $\hat{\mathbf{x}}_{t|t} = \mathbf{x}_{t|t-1}$
        \State $P_{t|t} = P_{t|t-1}$
    \EndIf

\EndWhile

\Statex
\State \textbf{Post-Disconnection Search Model}
\State \textit{(Scenario 1: Continued Navigation)}
\State $\mathbf{x}_{t+1} = F\mathbf{x}_t + G\mathbf{u}_t$
\State \textit{(Scenario 2: Propulsion Failure)}
\State $\mathbf{x}_{t+1} = F\mathbf{x}_t + \mathbf{w}_t,\;\mathbf{w}_t\sim\mathcal{N}(0,Q)$

\end{algorithmic}
\end{algorithm}

\section{Simulation Results}
\label{sec:simulation}

We conducted simulations to evaluate the proposed localization, prediction, and search framework under consistent experimental conditions. Five buoys were fixed at coordinates $[-800, -200, 3]$, $[-200, -800, 0]$, $[-800, -1000, 0]$, $[0, 0, 0]$, and $[-500, -500, -500]$. A total of 1000 test points were uniformly distributed in 3D space, starting from $[-100, -100, -50]$ with intervals of $60~\mathrm{m}$ along the $x$- and $y$-axes and $30~\mathrm{m}$ along the $z$-axis. The sampling interval was $10~\mathrm{s}$ for all experiments. This configuration was used consistently across the TDOA localization, Kalman filter prediction, and search model evaluations.
\subsection{Time Difference of Arrival (TDOA)}

\begin{figure}[!htbp]
    \centering
    \includegraphics[width=0.75\linewidth]{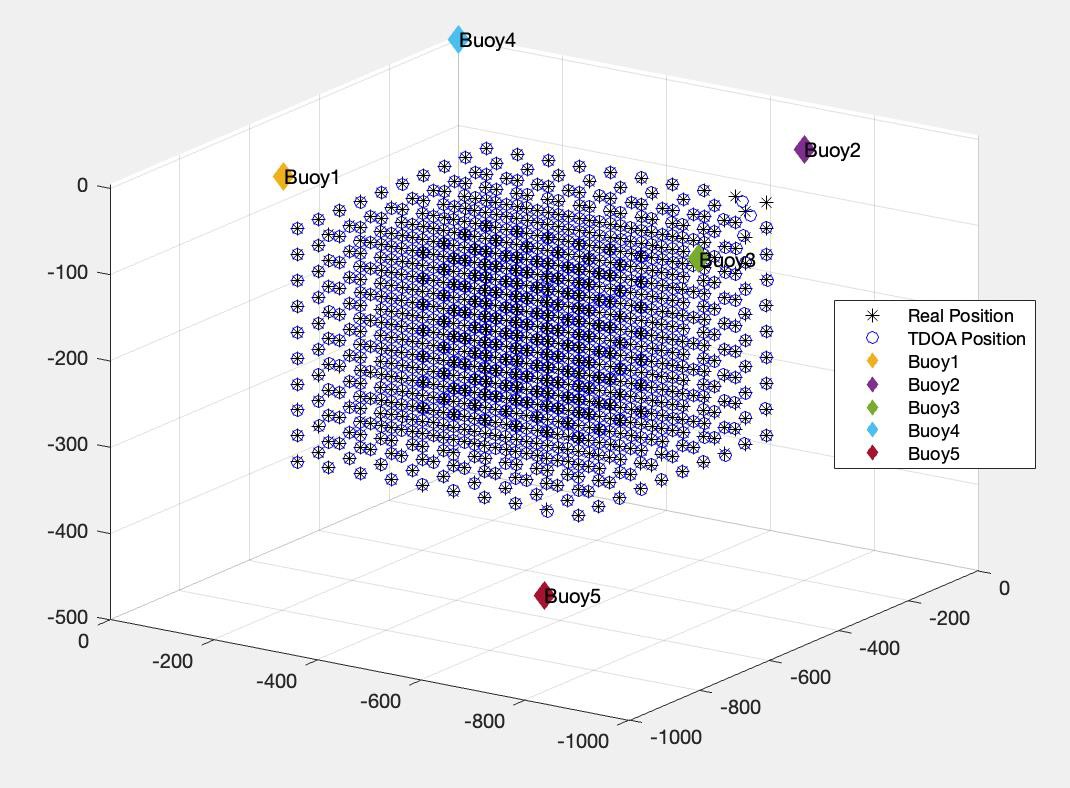}
    \caption{Localization results for 1000 test points. Blue circles denote true positions; yellow markers represent TDOA estimates.}
    \label{fig:tdoa_results}
\end{figure}

Figure~\ref{fig:tdoa_results} shows that the TDOA algorithm produces position estimates closely aligned with ground truth across all test points, demonstrating high localization fidelity under the given buoy configuration.

\begin{figure}[!htbp]
    \centering
    \includegraphics[width=0.75\linewidth]{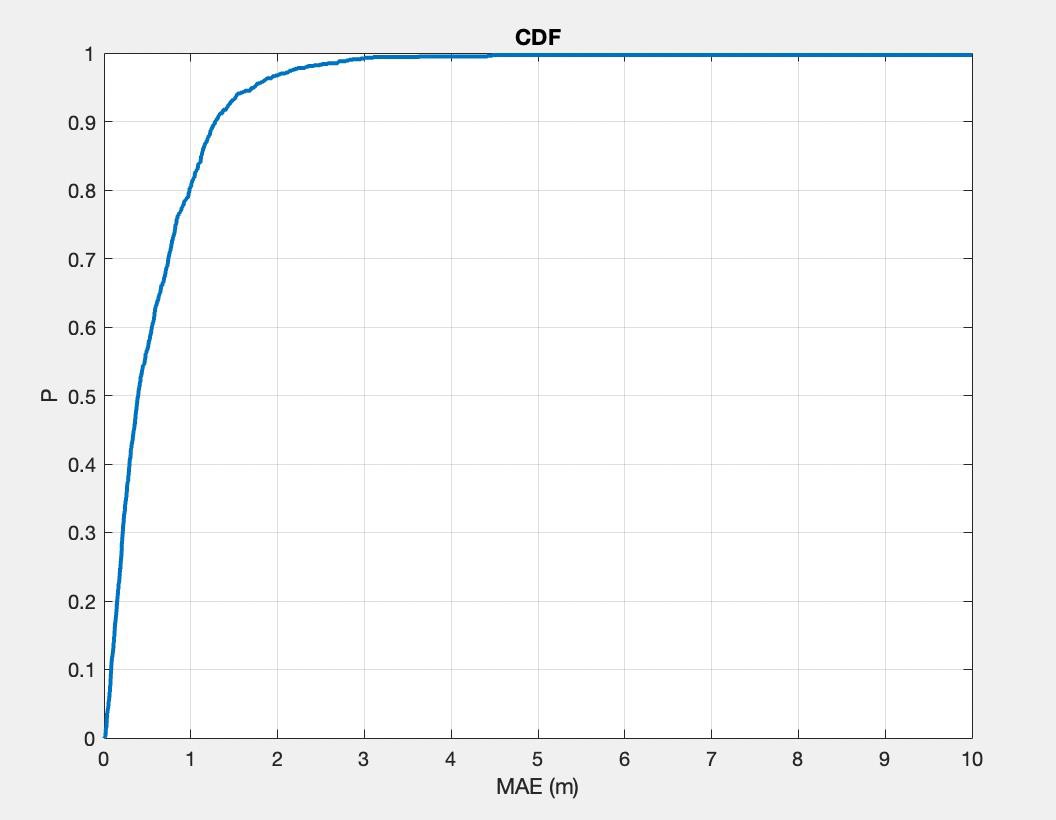}
    \caption{Cumulative Distribution Function (CDF) of Mean Absolute Error (MAE) for Submarine Localization. The CDF curve illustrates the probability that the localization error is less than or equal to a given MAE threshold. Nearly all test points have errors below 4~m, demonstrating high localization accuracy.}
    \label{fig:cdf_mae}
\end{figure}

Figure~\ref{fig:cdf_mae} shows the Cumulative Distribution Function (CDF) of the Mean Absolute Error (MAE) between the ground‑truth and TDOA‑estimated positions. The CDF begins at zero and approaches one, representing the probability that the localization error is below a given threshold. Nearly all test points exhibit MAE under $4\,\mathrm{m}$, indicating that the TDOA-based localization closely matches the true positions and provides reliable accuracy under the simulated buoy configuration.

The resulting position estimates serve as the measurement input for subsequent trajectory prediction via Kalman filtering, forming the foundation for the integrated localization–prediction pipeline evaluated in the following section.

\subsection{Kalman Filter Prediction}

\begin{figure}[!htbp]
    \centering
    \includegraphics[width=0.75\linewidth]{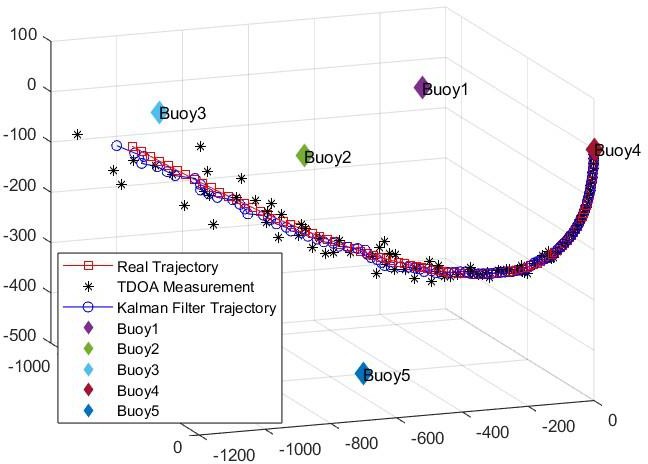}
    \caption{Simulation result of the prediction model. The real trajectory, TDOA measurements, and Kalman filter predictions are shown together for comparison.}
    \label{fig:kalman_prediction}
\end{figure}

Building on the verified localization accuracy, we next assess the Kalman filter’s ability to refine and predict the AUV’s motion in the presence of measurement noise. As shown in Figure~\ref{fig:kalman_prediction}, the Kalman filter produces a smooth and consistent trajectory, effectively suppressing the fluctuations observed in the raw TDOA measurements. This demonstrates that the filter successfully fuses sequential observations with the underlying motion model to generate reliable state estimates.

As time progresses, raw TDOA estimates drift further from the true trajectory, whereas the Kalman filter remains well aligned with ground truth by suppressing accumulated noise and uncertainty.

\begin{figure}[!htbp]
    \centering
    \includegraphics[width=0.75\linewidth]{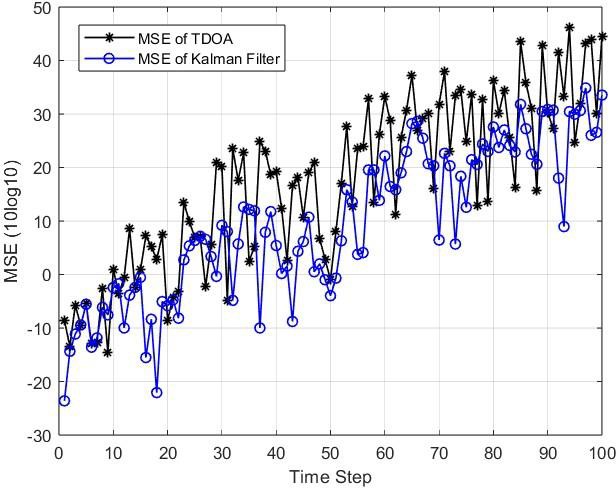}
    \caption{MSE comparison between the TDOA model and Kalman filter prediction. The Kalman filter achieves consistently lower error across time steps.}
    \label{fig:mse_comparison}
\end{figure}

Figure 7 provides a quantitative comparison of mean squared error (MSE) for both TDOA-only and Kalman-filtered predictions. The Kalman filter consistently achieves lower error across all time steps, demonstrating its robustness against observation noise and its superior trajectory prediction performance.

\subsection{Search Model under Communication Loss}

Having validated the localization and prediction subsystems, we next evaluate the search and recovery model under communication-loss scenarios.

\begin{figure}[!htbp]
    \centering
    \includegraphics[width=0.75\linewidth]{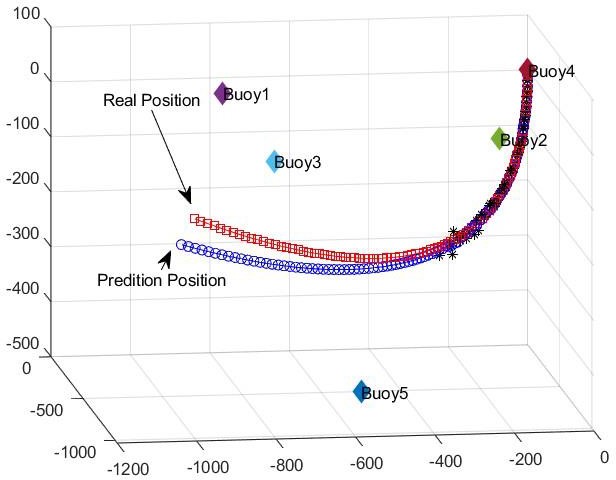}
    \caption{Simulation of the search process when disconnection occurs at $t=50~\mathrm{s}$. 
    The predicted trajectory (red) closely follows the real trajectory (blue) along the predetermined route.}
    \label{fig:search_t50}
\end{figure}

\begin{figure}[!htbp]
    \centering
    \includegraphics[width=0.75\linewidth]{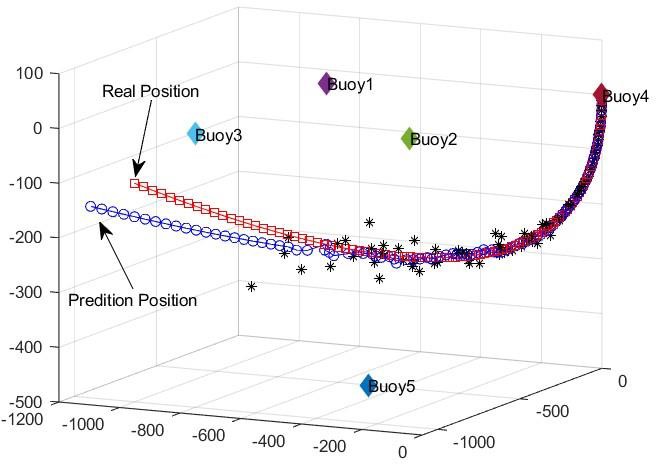}
    \caption{Simulation of the search process when disconnection occurs at $t=80~\mathrm{s}$. 
    The divergence between predicted and real positions increases with time as uncertainty accumulates.}
    \label{fig:search_t80}
\end{figure}
\noindent
Figures 8 and 9 illustrate the performance of the search model under communication loss at different time steps. For short-term disconnections (Figure~8), the Kalman-based prediction remains closely aligned with the actual submersible trajectory, demonstrating the framework’s ability to guide rescue efforts within a constrained search region. For longer disconnections (Figure~9), the predicted and true positions gradually diverge as uncertainty accumulates, yet the model continues to provide a principled estimate of the most probable search area. Although the discrepancy between predicted and true trajectories increases over time, the Kalman-based predictor continues to deliver reliable and practically meaningful position estimates. This capability supports targeted rescue operations even during extended communication loss, highlighting the framework’s resilience to real-world uncertainty. The simulation results demonstrate that the integrated framework enables adaptation of search strategies in response to evolving uncertainty, thereby enhancing the effectiveness of recovery operations.

\section{Conclusion and Future Work}
\label{sec:conclusion}
This paper presented an integrated framework for localization, prediction, and search for AUVs under uncertain communication conditions. Combining TDOA multilateration with Kalman filtering yields accurate localization and stable trajectory prediction. The search model captures how uncertainty grows after disconnection, enabling adaptive search‑region planning.

While the current framework assumes idealized signal conditions, real-world scenarios introduce additional challenges such as state noise, varying sound-speed profiles, and nonuniform buoy placement. 
Future work will incorporate adaptive calibration, stochastic current modeling, and noise estimation to improve the real-to-sim gap. 
Extending the framework to cooperative multi-AUV missions will further enhance coverage efficiency and resilience to signal loss.

We believe that researchers in underwater navigation, acoustic sensing, and predictive control will find this framework valuable, as it demonstrates a systematic way to address model uncertainty and enhance safety in real-world AUV operations.

\section*{Acknowledgment}

This work extends our submission to the 2024 Mathematical Contest in Modeling (MCM). We thank COMAP for organizing the competition and granting permission to further develop this research, as well as the MCM problem committee and judges for their valuable feedback.

\end{document}